\documentclass[10pt,twocolumn,letterpaper]{article}

\usepackage{times}
\usepackage{epsfig}
\usepackage{graphicx}
\usepackage{amsmath}
\usepackage{amssymb}
\usepackage{multicol}
\usepackage{balance}

\newcommand{\etal}{\textit{et al.}}

\begin{document}

\title{The Effect of Data Ordering in Image Classification}

\author{Ethem F. Can \\
\and
Aysu Ezen-Can \\
}

\maketitle

\begin{abstract}
The success stories from deep learning models increase every day spanning different tasks from image classification to natural language understanding. With the increasing popularity of these models, scientists spend more and more time finding the optimal parameters and best model architectures for their tasks. In this paper, we focus on the ingredient that feeds these machines: the data. We hypothesize that the data ordering affects how well a model performs. To that end, we conduct experiments on an image classification task using ImageNet dataset and show that some data orderings are better than others in terms of obtaining higher classification accuracies. Experimental results show that independent of model architecture, learning rate and batch size, ordering of the data significantly affects the outcome. We show these findings using different metrics: NDCG, accuracy @ 1 and accuracy @ 5. Our goal here is to show that not only parameters and model architectures but also the data ordering has a say in obtaining better results. 

\end{abstract}

\section{Introduction}

Deep learning has been taking over traditional machine learning methods on many tasks ranging from image classification to language modeling as they usually outperform their traditional counterparts. Even though deep learning has not been new to the researchers, with its re-birth in the Image-Net 2012 challenge, it started to gain a lot of attention such that now it is almost impossible to see an academic paper not containing at least a flavor of deep learning. 

After the AlexNet model~\cite{krizhevsky2012imagenet} winning the image classification task while reducing the error rate to its half of what was the state of the art back then, researchers started to employ deep learning methods to solve other computer vision tasks such as action recognition and object detection. 

The human movies database~\cite{Kuehne11} (HMDB) dataset was released as an action recognition dataset in 2011 and the creators of this dataset reported an accuracy of 20-25\%. It was around 40\% when ~\cite{can2013formulating} was published and jumped to 60\% with the early attempts of using convolutional neural networks (CNN) in action recognition~\cite{simonyan2014two}. Now the current state of the art on this particular dataset has exceeded 80\%~\cite{actionrecognitionrate}.

A similar pattern can be witnessed for the object detection task. VOC Pascal 2007 dataset was released for object detection and it even has its own challenge where teams of scientists compete to obtain the best scores. The non-deep learning methods were getting in the range of 20-40 mean average precision (mAP)~\cite{voc2007}. The involvement of deep learning in object detection helped raise this range to 70 mAP~\cite{zhao2019object} (these scores are obtained when only VOC 2007 data is used).

In the course of getting highly accurate results in many tasks, researchers also made adjustments in the ways to create the models. One major direction was to investigate models that require less number of parameters to learn. The VGG model~\cite{simonyan2014very} has more than 100 million parameters. Although it obtained respectable results in the Image-Net challenge, researchers moved to alternatives as training such a model requires a lot of resources. In addition to the requirement of extensive amount of resources, training a well converged VGG model (and also other very deep models) is usually difficult as gradients tend to vanish more when the network is deeper. A number of ways are proposed to tackle the problem such as auxiliary loss in a middle layer~\cite{szegedy2015going}. However, we can argue that the ``identity bypass connection" from He \etal~\cite{he2016deep} (a.k.a ResNet) help resolve the issue more than any previous attempts. The same authors then made some improvements to their methods to be able to train very very deep neural networks~\cite{he2016identity}. We should also note that ResNet includes another trick; batch-normalization: a method to standardize (using mean and variance) each batch of images.

The concept of non-sequential connections were studied before the ResNet models~\cite{hochreiter1997long,srivastava2015training}. Of course, ResNet models did not stop researchers to focus on the concept. Huang \etal~\cite{huang2017densely} proposed the idea of connecting layers with each other. They named this concept as DenseNet where output of a layer has connections from itself as well as previous layers, therefore the network is densely connected. 

Although non-sequential shortcut connections in ResNet and DenseNet give researchers a break, training a very deep network still would take a good amount of time. Dropout, an approach to prevent over-fitting via ignoring some neurons randomly during training~\cite{srivastava14a}, usually yields less training time for each epoch. On a somehow similar motivation, \cite{huang2016deep} studied dropping out layers during training and \cite{veit16} experimented dropping out layers in testing as well. They argue that some of the layers would be redundant and can be removed without a performance degrade. Keeping this information in mind, searching a network on a large search space, consisting of a set of layers, would be the natural next step. 

The selling point of deep learning methodologies in their early days was to highlight that they remove the need for feature engineering to solve problems. It was a good marketing point until researchers realized that they now need to engineer the hyper-parameters (e.g., learning rate and even the architecture itself) of neural network architectures rather than features. That said, the results were so pleasantly convincing, solutions to deal with hyper-parameter search have been proposed.  

Searching for the optimal hyper-parameters such as learning rate are considered to be hyper-parameter tuning, while finding an optimal architecture, also sometimes called model design, is called neural architecture search (NAS). An example of a NAS study is from Zaph and Le~\cite{zoph2016neural}. They focus on a recurrent neural network to generate the model descriptions while training it using recurrent neural networks (RNN) as a controller to decide if a child network is an optimal one or not~\cite{zoph2016neural}. There were previous studies including but not limited to the ones from Bergstra~\etal\cite{bergstra2012}, Bergsta~\etal\cite{bergstra2013}, and Mendoza~\etal\cite{mendoza2016towards}. However, Zoph and Le~\cite{zoph2016neural} claimed that their study is more general and more flexible than any previous study on NAS. The authors compared their results with the state-of-the-art results for the image classification as well as language modeling tasks. However, it was hard to see significant, or sometimes any, improvements over the state-of-the-art results.

CIFAR10 and MNIST are the first choices that scientists target while conducting experiments to show the effectiveness of their method. These datasets are very well studied but they show no proof of scalability. Zoph~\etal\cite{zoph2018learning} made exactly the same point. They perform their experiments on the ImageNet dataset along with the COCO dataset. Actually, they achieved pseudo-scalability as they again search an optimal architecture on a proxy dataset such as CIFAR10 and then transfer it to larger datasets such as ImageNet. Even though it is pseudo, they achieved good results on these datasets compared to the state-of-the-art results.

One of the drawbacks about claiming scalability is that the term itself is being subjective. For example, a search algorithm to find an optimal solution requiring four days with 500 P100 GPUs would be counted as scalable at some part of the world, whereas the same setting would not be accepted as scalable if you have only one GPU where the same task would be completed in 2,000 days. Liu~\etal\cite{liu2018progressive} proposed a solution where they start with searching for a good convolutional ``cell" rather than a full architecture. Each cell consists of a number of convolutional blocks. A final output is determined by the size of the training set and the desired running time and it basically is a number of stacked cells. The results clearly show that progressive NAS (PNAS) speeds up search compared to NAS while preserving a similar level of effectiveness. Zhou~\etal\cite{zhou2019bayesnas} also addressed the inefficient search methods problem and proposed a Bayesian approach for NAS that also includes network compression. We should also note that some of the best results on the ImageNet dataset are obtained with architectures designed using NAS methods. This would be a reason, among many others, architecture search has been very popular especially for the last a couple of years~\cite{litnas,elsken2019neural}.

Up to now, we have discussed how deep learning takes over conventional machine learning algorithms along with multiple ways to make deep learning models efficient and effective. The approaches to make the models efficient and effective operates in the model itself. We have not discussed the operations that can be applied to the data. One of the major data operations is data augmentation techniques such as random cropping. Even though these data techniques help boosting the final results, we would like to focus on a different stream: data ordering.

GPUs are heavily used to train deep learning models as matrix calculations are quite fast in such hardware. Given that deep learning methods include, or converted into, a lot of matrix operations, using GPUs enable us to speed up the training and inference phases of deep learning models. Stochastic gradient descent (SGD) is the major optimization technique used in deep learning methods. SGD does not require loading the entire training set at once but it works in small batches. This enables us to use affordable and existing GPU hardware during training. The gradients are updated and the error is propagated at batch level rather than at the epoch level (e.g., traversing the entire training set). The data is selected randomly for each batch and more importantly it should be shuffled after completing an epoch. 

Most of the solutions we have discussed so far use the entire or even an up-sampled and larger version of the training data. Efforts to achieve efficient and ideally effective models include skipping some of the layers, dropping some layers, and searching a small network in a search space. This shows that some redundant work might be performed during training as removing, pruning, or selecting a lighter model would yield similar or better results. Keeping this motivation in mind, we hypothesize that crafting the training data would also help us efficiently train networks while preserving the effectiveness. In other words, we hypothesize that \textit{there exists such a data ordering for the batches that yield better results compared to other orderings}. To support this hypothesis, we here split the work into two folds: 1) showing that some ordering would be better than other orderings and 2) finding out that particular ordering. Proving the latter one is quite difficult. One of the reasons is that we should find a way to select a particular ordering that consistently outperforms its counterparts in most of the tasks and with different parameters. There are many moving parts and each of which needs to be controlled to draw a conclusion in the direction we are targeting. Rather than tackling these two folds at once, in this paper, we focus on showing \textit{that different ordering of data in batches would change the effectiveness of a model}, which is showing the first fold. In order to show this, we focus on different data orderings in batches while controlling the learning rate, batch size, and model type on the image classification task using the ImageNet dataset. We prefer using the ImageNet dataset as working on toy datasets such as CIFAR10 or MNIST would deceive us while we are drawing a conclusion due to their size. Controlling these parameters enable us to show that a certain data ordering in batches provide better classification accuracies independent of their different values. Having a consistent winning ordering with different learning rates, different batch sizes, and different architectures suggests us that better batches would be created. Of course the critical question we are leaving for the future work is to come up with an algorithm to tailor the data in batches that yield better classification accuracies. 

\section{Experiments and Discussion}

We dedicate this paper to investigate the effect of data ordering to the model effectiveness. Therefore, we conducted experiments with different data orderings. To be able to tell if the model accuracies are changing because of data ordering or because of other parameters such as learning rate, we experimented with various different parameter combinations. We used two well-known model architectures: DenseNet and ResNet. We picked a lighter version of ResNet (i.e., ResNet18) to experiment our goal on a network that has residual connection while providing reasonable image classification results on the ImageNet dataset.  We also focused on another network that has more dense connections and provides very good classification results (DenseNet). 

\begin{figure}[!]
\centering
\includegraphics[width=0.48\textwidth]{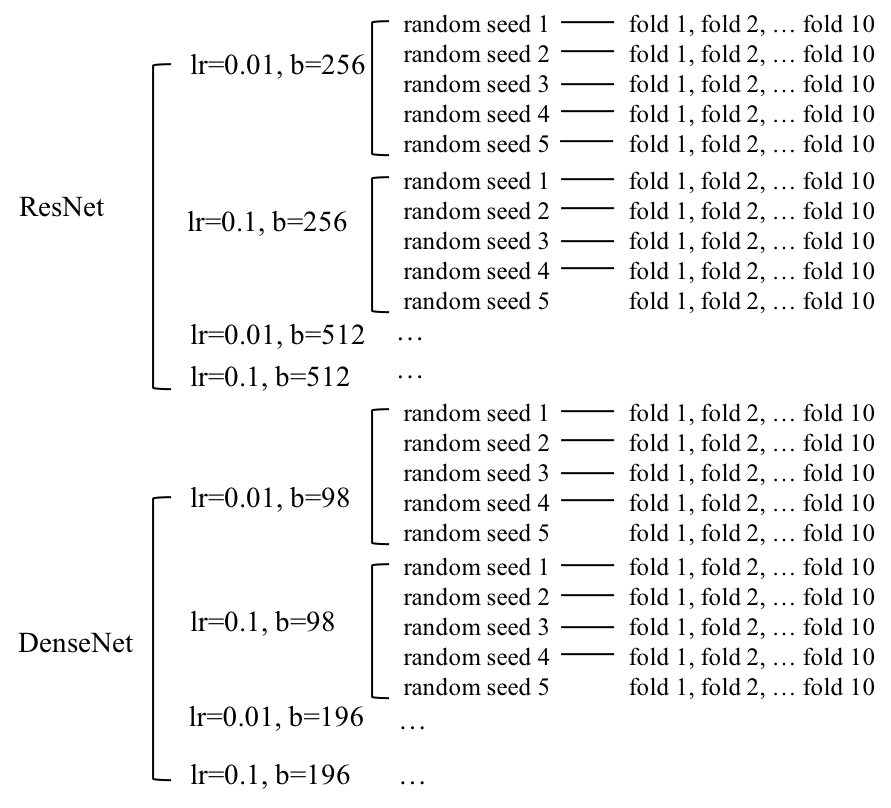}
\caption{Models and parameters used in experiments. }
 \label{fig:experiments}
\end{figure}

For each model architecture, we conducted four set of experiments (i.e., two different learning rates and two different batch sizes). For each set we ran the experiments with five different random seeds ran for one epoch, to eliminate the effect of different initializations. On top of these settings, we created 10 random data orderings (we call them folds) to show the results are not due to a certain ordering but are consistent. At the end of these experiments, we obtained a long list of experimental results therefore we cannot present all of the folds in this paper. However, running this many experiments helped us conclude the importance of data ordering while eliminating the effect of different parameters including model architectures, random seeds, learning rates, and batch sizes. The experiments conducted are shown in Figure  \ref{fig:experiments}.

\begin{figure}[!h]
\includegraphics[width=0.48\textwidth]{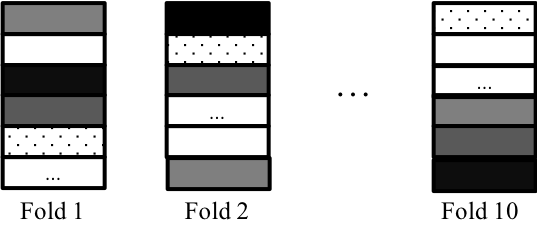}\par
\caption{Visualization representing data shuffling in folds.}
\label{folds}
\end{figure}

\begin{figure*}[!h]
\begin{multicols}{2}
\includegraphics[width=0.48\textwidth]{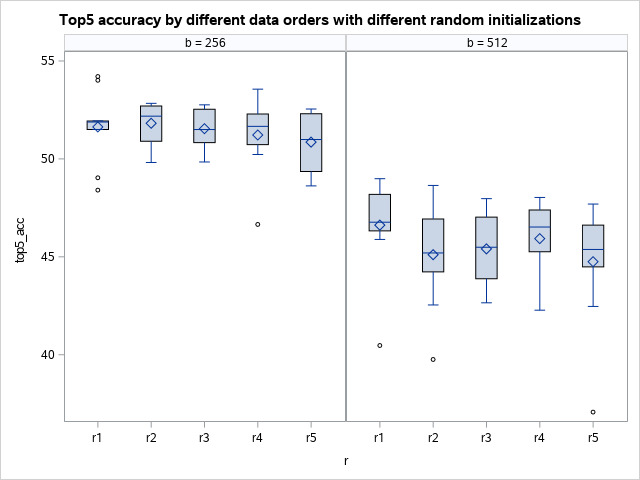}\par 
\includegraphics[width=0.48\textwidth]{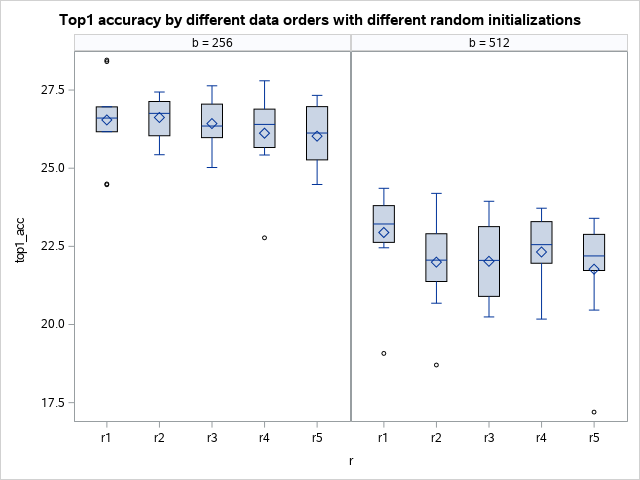}\par 
    \end{multicols}
\begin{multicols}{2}
\includegraphics[width=0.48\textwidth]{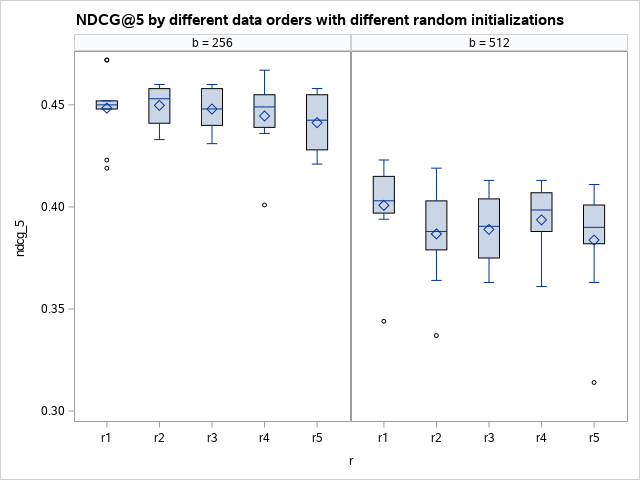}\par
\includegraphics[width=0.48\textwidth]{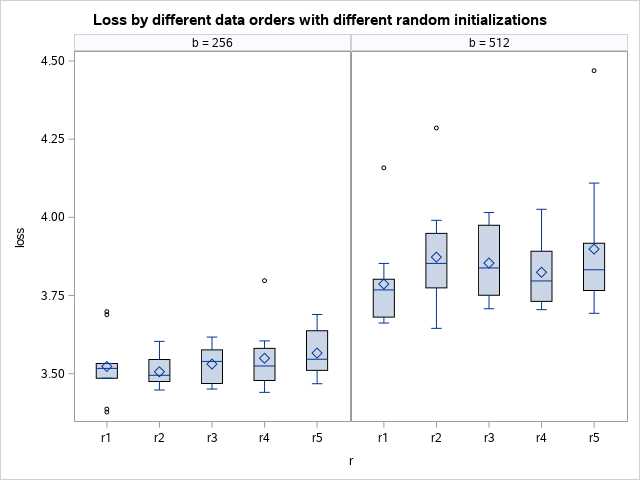}\par
\end{multicols}
\caption{Top5 accuracy, Top1 accuracy, and NDCG@5 results as well as loss values when using the ResNet18 model with lr=0.1. For each figure, the box on the left shows results with batch size=256 and the box on the right shows batch size = 512. r1 through r5 show different random seeds. }
\label{r18_lr01_results}
\end{figure*}

We visualize the classification results for ResNet18 with different evaluation metrics (accuracy @ 5, accuracy @ 1, and NDCG@5) as well as the loss value using learning rate = 0.1 in Figure~\ref{r18_lr01_results}. We included normalized discounted cumulative gain (NDCG@5) because we believe that it is a better metric than top5 accuracy as it differs if an accurate classification is in the first place or the fifth place.

Each box plot shows the distribution of results for different folds. Since each fold has a different data shuffling, having a wide variety in the box plot means that different data ordering gave different results. Different folds provide a wide range of classification results even with different initializations. This shows that independent of the initialization, we obtain very different classification results with different data orderings. In other words, there exists a data ordering that provides better results compared to another data ordering. This is pretty much the main goal of this paper, while keeping the ultimate goal of finding that order. That said, we limit the scope of this paper to empirically prove the existence of such orderings and leave the ultimate goal as future work.

In addition to different initializations, we also tested against different batch sizes and learning rates. Independent of these changing parameters, there exists such a data order that provides better results compared to another one. The experimental results with a different learning rate (i.e., lr=0.01) align with our findings where learning rate is 0.1. Further, even though we get more wider boxes in the plots with a larger batch size, it is still clear that we get better classification results with some data orderings than others.

\begin{figure*}
\begin{multicols}{2}
\includegraphics[width=0.48\textwidth]{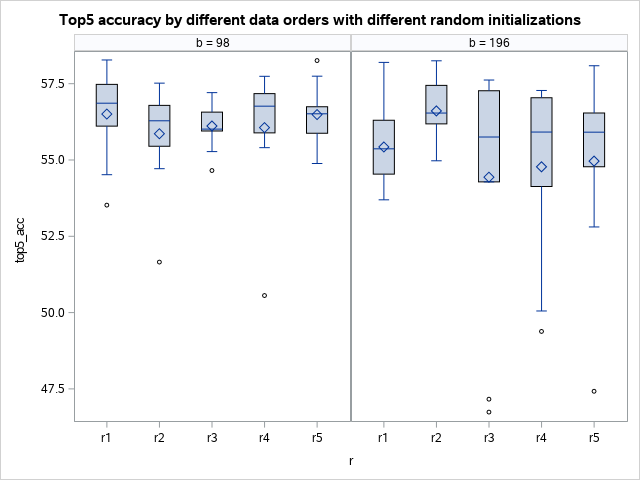}\par 
\includegraphics[width=0.48\textwidth]{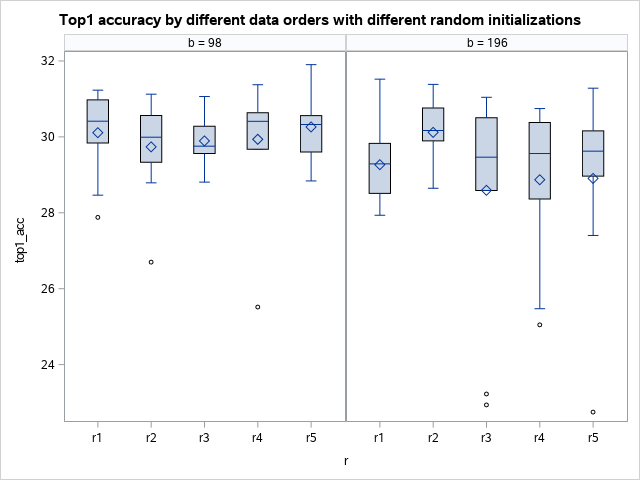}\par 
    \end{multicols}
\begin{multicols}{2}
\includegraphics[width=0.48\textwidth]{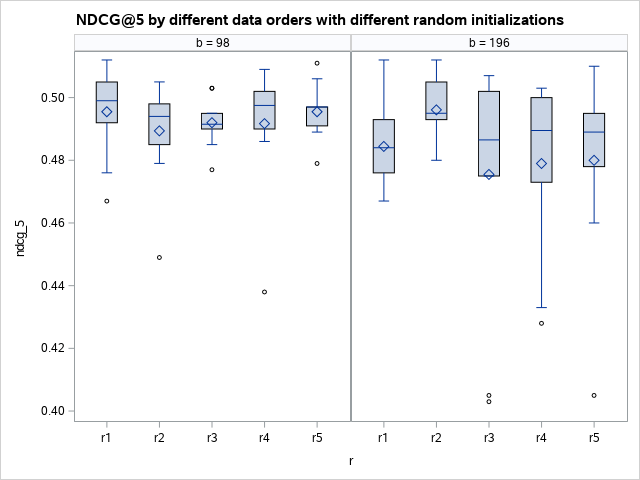}\par
\includegraphics[width=0.48\textwidth]{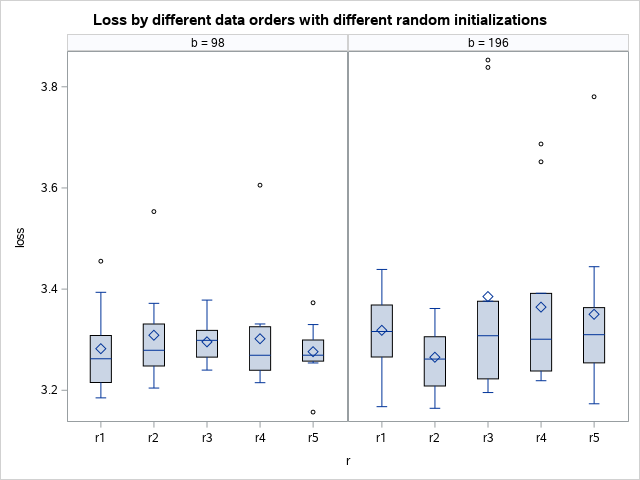}\par
\end{multicols}
\caption{Top5 accuracy, Top1 accuracy, and NDCG@5 results as well as loss values when using the DenseNet model with lr=0.1. For each figure, the box on the left shows results with batch size=98 and the box on the right shows batch size = 196. r1 through r5 show results with different random seeds. }
\label{d121_lr01_results}
\end{figure*}

So far, we have discussed the results that we obtained using the ResNet18 network. We also evaluate our hypothesis using the DenseNet (121 layer) network to show that our findings are not dependent upon a certain model. We provide the results that we obtained using the DenseNet architecture in Figure~\ref{d121_lr01_results}. The results are pretty much aligning with the results we obtained using the ResNet18 architecture. The boxes in the plots are a little bit wider in the DenseNet case. Not surprisingly the accuracies are higher with DenseNet compared to ResNet18. This might be a reason of the wider boxes, which also means a wider range of results obtained with different data orderings. Note that in the DenseNet case we were able to fit 196 images per batch, while we were able to fit 512 images in a batch with ResNet18. We perform all the experiments using a V100 card with 32GB of memory. 

\begin{table*}
\begin{center}
 \begin{tabular}{|p{4.5cm}|p{2.5cm}|p{2.5cm}|p{2.5cm}|p{2.5cm}|p{2.5cm}|} 
\hline
\textbf{Model \& parameters} &  \textbf{NDCG \textit{p-value}} & \textbf{acc@1 \textit{p-value}} & \textbf{acc@5 \textit{p-value}} & \textbf{\# of significant folds}  \\ 
 \hline
 ResNet18: lr=0.1,	b=512 &0.0005 & 0.0013&0.0003 & 16 \\
 ResNet18: lr=0.01	, b=512& 0.004& 0.0003&0.005 &  14\\
 DenseNet121: lr=0.01,	b=196  &0.0017 & 0.0007 &0.002 &12 \\
 DenseNet121: lr=0.01, b=98 & 0.1363& 0.0318 & 0.0203& 11\\
 \hline 
\end{tabular}
\end{center}
  \caption{ANOVA results comparing different experimental settings. To show that data ordering affects the model performance, we expected at least one fold to be significantly different from other folds.}
 \label{table:anova}
\end{table*}

In addition to the experiments with a learning rate of 0.1, we also conducted experiments where the learning rate is set to 0.01 using the DenseNet architecture. Similar to the ResNet findings, the results show that different data orderings provide different results. Note that, the differences are larger when we use a learning rate of 0.1 compared to a learning rate of 0.01. We observed a similar pattern with the ResNet architecture.

To investigate whether or not the differences in results are significant or not, we conducted ANOVA tests. For each group of experiments, we performed one-way ANOVA tests. For each test, we have one architecture, learning rate and batch-size set. There are 10 series (10 folds) with 5 elements in each (5 random initializations). We hypothesized that there is at least one fold that is significantly different from the rest of the folds in each CNN architecture. By showing this, we can conclude that there exists a random initialization that creates significantly different results from the rest of the data orderings.

The ANOVA tests revealed that data ordering significantly changes the model performance. Table \ref{table:anova} shows \textit{p} values obtained from ANOVA tests and the number of significant folds. The number of significant folds are calculated by comparing each fold with other folds and counting the number of significantly different fold pairs. As can be seen in the table, the models have more than 1 significantly different fold. This shows that several data orderings create different results.

Experimental results show that data ordering has an impact on effectiveness of a model no matter what the learning rate/ batch size / model architecture are. We have seen that for some learning rate and batch size combinations, the difference the data ordering creates is even bigger. 

\begin{figure*}[t]
\begin{multicols}{2}
\includegraphics[width=0.48\textwidth]{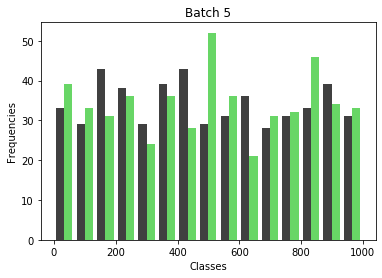}\par 
\includegraphics[width=0.48\textwidth]{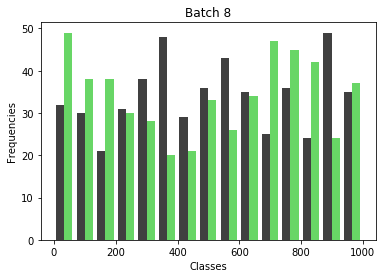}\par 
    \end{multicols}
\begin{multicols}{2}
\includegraphics[width=0.48\textwidth]{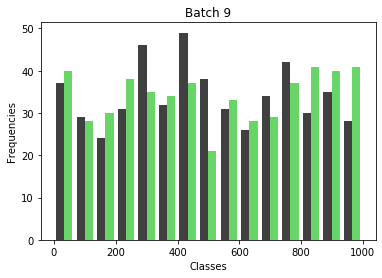}\par
\includegraphics[width=0.48\textwidth]{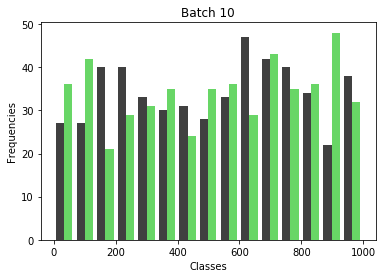}\par
\end{multicols}
\caption{Histograms of classes in different batches for ResNet18 model with batch size = 512.}
\label{batches}
\end{figure*}

So far, we have discussed the effects of data ordering and presented experimental results to support our hypothesis. After these encouraging findings, we investigated the bathes of folds to see if there is any clear signs to trigger these differences. We looked at the class distributions (most frequent and least frequent) in the batches.  Histograms of class distributions in some of the batches comparing two different folds (fold 1 and fold 8) are shown in Figure~\ref{batches}. As can be seen in the figure, the number of observations from each class are different in different folds. For example, in batch 5, fold 1 (green bar) has more observations from classes 500s than fold 8 (gray bar). Similarly, in batch 8, the number of observations with classes 300s are much more in fold 8 than fold 1. These findings help justify our hypothesis that the batches indeed have different distributions of data with different number of observations from each class. Batch 10 also shows differences in the distribution of classes. In batch 10, there are 236 classes represented in fold 8 but not in fold 1. There are 240 classes represented in fold 1 but not in fold 8 in this batch. When a class is not represented in a batch, the training for that iteration get affected, therefore causing the differences in the results. The number of unrepresented classes are similar in other batches as well. For example, batch 9 has 237 unrepresented classes in fold 8 but not in fold 1 and 254 unrepresented classes in fold 1 that are present in fold 8.

Perhaps our analysis is limited to the visible signs, and we believe that further investigations in this direction would yield results. Maybe, the perfect data ordering would be revealed by training another model. Definitely, we will be working on more formal ways of identifying better data orderings, given that we now have strong evidence that there exists a data ordering that would perform better than others.

\section{Conclusion}

Deep learning models have achieved state-of-the-art results in many tasks from image classification to natural language understanding. These improvements led to more scientists spending time on tuning parameters and searching for the optimal architectures for their tasks. In this paper, we showed another important perspective for obtaining good results: data ordering. 

Focusing on image classification task, we ran experiments with different learning rates and batch sizes while using different data orderings to the models (i.e. shuffled data). In terms of evaluating effectiveness of models, we used several different metrics: NDCG@5, ACC@1 and ACC@5. For all of these metrics, the conclusion was the same: data ordering changes effectiveness of models significantly and therefore it is an important aspect to be taken into account while building highly effective deep learning models. ANOVA analyses of the experimental results confirmed that the folds are significantly different from each other, verifying our hypothesis. 

Given the importance of data ordering, we believe that it is important for researchers to provide the order of observations when reporting  results. Even after using the same code, same dataset, same learning rate and batch size, it is possible to obtain different results by two researchers if the ordering of the data is not the same. Therefore, we invite the research community to share the functions they use for ordering their datasets to increase reproducibility of the results. 

\balance
\clearpage






\newpage
{\small
\bibliographystyle{ieee}
\bibliography{egbib}
}
\balance

\end{document}